%% file: main.tex
\documentclass[10pt,twocolumn,letterpaper]{article}

\usepackage[pagenumbers]{cvpr} 

\input{preamble}
\definecolor{cvprblue}{rgb}{0.21,0.49,0.74}
\definecolor{subjectcolor}{rgb}{0.42, 0.11, 0.85}
\definecolor{objectcolor}{rgb}{0.85, 0.55, 0.11}
\definecolor{verbcolor}{rgb}{0.54, 0.87, 0.37}
\usepackage[pagebackref,breaklinks,colorlinks,allcolors=cvprblue]{hyperref}
\usepackage{soul,xcolor}
\def\paperID{8110} 
\def\confName{CVPR}
\def\confYear{2026}

\title{Click2Graph: Interactive Panoptic Video Scene Graphs from a Single Click}

\author{Raphael Ruschel*\\
UC Santa Barbara\\
Electrical \& Computer Engineering\\
{\tt\small raphael251@ucsb.edu}
\and
Hardikkumar Prajapati*\\
UC Santa Barbara\\
Electrical \& Computer Engineering\\
{\tt\small hprajapati@ucsb.edu}
\and
Md Awsafur Rahman\\
UC Santa Barbara\\
Electrical \& Computer Engineering\\
{\tt\small awsaf@ucsb.edu}
\and
B. S. Manjunath\\
UC Santa Barbara\\
Electrical \& Computer Engineering\\
{\tt\small manj@ucsb.edu}
}

\begin{document}
\newcommand{\colorTrip}[3]{%
  \textcolor{subjectcolor}{\textbf{#1}}, %
  \textcolor{objectcolor}{\textbf{#2}}, %
  \textcolor{verbcolor}{\textbf{#3}}%
}

\maketitle
\input{sec/0_abstract}
\input{sec/1_intro}
\input{sec/2_related}
\input{sec/3_method}
\input{sec/4_dataset}
\input{sec/5_evalmetrics}
\input{sec/6_result}

\input{sec/7_Conc_future}
{
    \small
    \bibliographystyle{ieeenat_fullname}
    \bibliography{refs/hoi, refs/generic, refs/prompt_hoi, refs/vrl}
}


\end{document}

%% file: sec/0_abstract.tex
\begin{abstract}
State-of-the-art Video Scene Graph Generation (VSGG) systems provide structured visual understanding but operate as closed, feed-forward pipelines with no ability to incorporate human guidance. In contrast, promptable segmentation models such as SAM2 enable precise user interaction but lack semantic or relational reasoning. We introduce \textbf{Click2Graph}, the first interactive framework for \textit{Panoptic Video Scene Graph Generation (PVSG)} that unifies visual prompting with spatial, temporal, and semantic understanding. From a single user cue, such as a click or bounding box, Click2Graph segments and tracks the subject across time, autonomously discovers interacting objects, and predicts $\langle subject, object, predicate\rangle$ triplets to form a temporally consistent scene graph. Our framework introduces two key components: a \textbf{Dynamic Interaction Discovery Module} that generates subject-conditioned object prompts, and a \textbf{Semantic Classification Head} that performs joint entity and predicate reasoning. Experiments on the OpenPVSG benchmark demonstrate that Click2Graph establishes a strong foundation for user-guided PVSG, showing how human prompting can be combined with panoptic grounding and relational inference to enable controllable and interpretable video scene understanding.
\end{abstract}

%% file: sec/1_intro.tex
\section{Introduction}

Understanding not only \emph{what} appears in a video but \emph{how entities interact} is a core challenge in intelligent perception. This capability is desired in applications in robotics, autonomous agents, assistive systems, and surveillance, where downstream decisions depend on correctly interpreting actions, intentions, and relationships. Scene Graph Generation (SGG) has emerged as a powerful representation for such structured understanding, evolving from static image reasoning~\cite{xu2017scenegraph, zellers2018neural} to dynamic, video-based formulations that capture temporal context~\cite{cong2021spatial, khandelwal2022iterative}. More recently, panoptic scene graph generation has advanced grounding fidelity by replacing bounding boxes with pixel-level masks~\cite{yang2022panoptic, yang2023panoptic}, enabling fine-grained grounding of classes, especially for objects with irregular shapes (commonly refered as "stuff" classes), such as floor and sky.

\begin{figure*}
    \centering
    \includegraphics[width=1.0\linewidth]{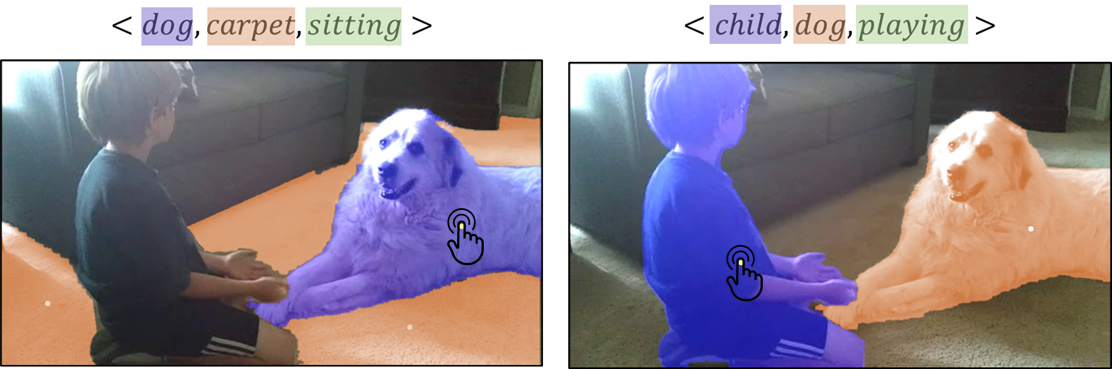}
    \caption{On the left example, the user clicked on the \textcolor{subjectcolor}{$\langle \text{\textbf{dog}} \rangle$}, and Click2Graph segmented the \textcolor{objectcolor}{$\langle \text{\textbf{carpet}} \rangle$} and predicted the \textcolor{verbcolor}{$\langle \text{\textbf{sitting}} \rangle$} activity. On the right, we have a prompt on \textcolor{subjectcolor}{$\langle \text{\textbf{child}} \rangle$} which yields \textcolor{objectcolor}{$\langle \text{\textbf{dog}} \rangle$}, \textcolor{verbcolor}{$\langle \text{\textbf{playing}} \rangle$} as associated object and activity.}
    \label{fig:pipeline_sample}
\end{figure*}

Despite these advances, existing video SGG and PVSG pipelines remain fully automated and closed-loop. Once a model overlooks an occluded object, misclassifies a rare interaction, or drifts during tracking, the user has no mechanism to intervene. This lack of controllability is problematic in complex or safety-critical environments, where correcting errors or directing model attention is essential. At the same time, a new class of promptable segmentation models, most notably SAM and SAM2~\cite{kirillov2023segment, ravi2024sam}, has demonstrated the power of direct \emph{visual prompting}. With a simple click or box, users can obtain precise, temporally consistent segmentation masks. Yet these models are inherently class-agnostic and relation-agnostic: they determine \emph{where} objects are but not \emph{what} they are or \emph{how} they interact.

This disconnect reveals a fundamental gap: current PVSG systems lack user guidance, and current interactive segmentation models lack semantic structure. We address this gap with \textbf{Click2Graph}, the first framework for \emph{user-guided Panoptic Video Scene Graph Generation}. From a single visual cue, such as clicking a subject in any frame, Click2Graph:
\begin{enumerate}
    \item Segments and tracks the prompted subject across time,
    \item Autonomously discovers and segments interacting objects, and
    \item Predicts $\langle subject, object, predicate\rangle$ relationships to form a temporally consistent scene graph.
\end{enumerate}
Figure~\ref{fig:pipeline_sample} illustrates how distinct scene graphs can be produced depending on which entity the user prompts, highlighting the controllability of the system.

Click2Graph introduces two components that supply the missing semantic and relational reasoning. A \textbf{Dynamic Interaction Discovery Module (DIDM)} generates subject-conditioned object prompts, enabling automatic discovery of entities participating in interactions. A \textbf{Semantic Classification Head (SCH)} performs joint subject, object, and predicate inference over the discovered segments, producing structured scene graph outputs. Together, these components elevate promptable segmentation from geometric mask extraction to full panoptic video scene graph generation.

Our contributions are summarized as follows:
\begin{itemize}
    \item \textbf{User-Guided Panoptic Video Scene Graphs.} We introduce the first interactive PVSG framework that converts a single visual prompt into a temporally consistent panoptic scene graph, enabling controllable, interpretable video analysis.
    \item \textbf{Dynamic Interaction Discovery.} We propose a novel module that generates subject-conditioned prompts to discover interacting objects, naturally supporting multi-subject and multi-object reasoning.
    \item \textbf{Semantic Reasoning atop Promptable Segmentation.} A dedicated classification head predicts subject--object pairs labels and the relationship between them, bridging the gap between prompt-based segmentation and structured semantic inference.
\end{itemize}

Click2Graph establishes a new paradigm for video scene understanding by combining human guidance, pixel-level grounding, and relational reasoning within a unified architecture. As shown in our experiments on the OpenPVSG benchmark, this paradigm enables controllable and interpretable scene graph generation while offering a practical path toward real-world interactive video analytics.

%% file: sec/2_related.tex
\section{Related Works}

\begin{table*}[t]
\centering
\caption{A comparative analysis of Scene Graph Generation paradigms. Click2Graph is the first to unify video-level temporal reasoning, panoptic-level spatial precision, and user-guided visual prompting for end-to-end tracking and relationship prediction.}
\label{tab:comparison}
\resizebox{\textwidth}{!}{%
\begin{tabular}{l|ccccc}
\hline
\textbf{Method} & \textbf{Modality} & \textbf{Granularity} & \textbf{Interaction Type} & \textbf{End-to-End Tracking} & \textbf{Relationship Prediction} \\ \hline
Traditional SGG (e.g., MOTIFS \cite{zellers2018neural}) & Image & Box & None & N/A & Yes \\
Video SGG (e.g., STTran \cite{cong2021spatial}) & Video & Box & None  & Yes & Yes \\
Panoptic SGG (e.g., PSGFormer \cite{yang2022panoptic}) & Image & Mask & None  & N/A & Yes \\
Panoptic Video SGG (PVSG) \cite{yang2023panoptic} & Video & Mask & None  & Yes & Yes \\
Text-Prompted SGG (e.g., VLPrompt \cite{zhou2023vlprompt}) & Image & Mask & Text & N/A & Yes \\
\textbf{Click2Graph (Ours)} & \textbf{Video+Image} & \textbf{Mask} & \textbf{Visual (Click/Box)} & \textbf{Yes} & \textbf{Yes} \\ \hline
\end{tabular}%
}
\end{table*}

Our work lies at the intersection of video scene graph generation, panoptic-level scene understanding, and interactive visual analysis. Below, we position Click2Graph within each of these domains. Table \ref{tab:comparison} shows a summary of the related domains.

\subsection{Advances in Video Scene Graph Generation}

Scene Graph Generation (SGG) was first developed for static images~\cite{xu2017scenegraph, zellers2018neural}, and later extended to videos (VidSGG) to capture temporal dynamics~\cite{khandelwal2022iterative, STTran-TPI, li2022dynamic, VsCGG, Lin2024TD2NetTD, feng2023exploiting, wang2024oed, zhang2023end}. Transformer-based approaches such as STTran~\cite{cong2021spatial}, DDS~\cite{iftekhar2023ddsdecoupleddynamicscenegraph}, and VSG-Net~\cite{ulutan2020vsgnet} improved long-range temporal reasoning and robustness to clutter. Another thread of work addresses the heavy long-tail distribution of predicates through debiasing methods such as TEMPURA~\cite{nag2023unbiased} and VISA~\cite{li2024unbiased}, while DiffVSGG~\cite{chen2025diffvsgg} frames video SGG as an iterative denoising problem.

Although these methods advance automated scene graph reasoning, they operate as closed-loop systems: once the model misdetects or misclassifies an entity, the user cannot intervene. Click2Graph introduces this missing interactive dimension, enabling subject-specific, user-directed scene graph construction.

\subsection{Panoptic-Level Scene Understanding}

To improve spatial precision, recent works replace bounding boxes with pixel-level masks. Panoptic Scene Graph Generation (PSG)~\cite{yang2022panoptic} grounds all entities, including “stuff’’ classes, in panoptic masks. This paradigm was extended temporally in the Panoptic Video Scene Graph (PVSG) task~\cite{yang2023panoptic}, which provides temporally consistent panoptic annotations through the OpenPVSG benchmark.

Click2Graph builds on this foundation but differs from prior PVSG approaches by introducing user control. Instead of producing a full-frame graph in a fully automated manner, we allow a user to specify a subject of interest and generate an interaction-centric scene graph guided by that prompt.

\subsection{Promptable and Interactive Scene Analysis}

Interactive reasoning has emerged in adjacent domains but remains underexplored for scene graph generation. Existing approaches fall into two categories: text-prompted and visually-guided methods.

\subsubsection{Text-Prompted Generation}

Several SGG methods incorporate language guidance. Ov-SGG~\cite{he2022towards} and CaCao~\cite{cacao-sgg} use text prompts for open-vocabulary predicate detection, while VLPrompt~\cite{zhou2023vlprompt} integrates LLM-derived priors to improve panoptic SGG. Although language prompts provide rich semantics, they lack spatial specificity, text cannot uniquely and precisely ground pixel-level subjects. These systems also depend on language availability and may not generalize across settings.

In contrast, Click2Graph uses direct visual prompts (points, boxes or masks), which are universal, unambiguous, and spatially precise.

\subsubsection{Visually-Guided Interaction}

Visually guided interaction remains largely unexplored for scene graph generation. Prior work has examined interactive image or 3D scene graph editing~\cite{ashual2020interactive, mittal2019interactive, li2025sgsg}, and interactive video object segmentation (VOS) allows tracking of a single prompted object~\cite{heo2021guided}. However, these methods lack interaction discovery and semantic relationship reasoning.

To our knowledge, \textbf{Click2Graph is the first framework to leverage direct visual prompts for end-to-end Panoptic Video Scene Graph Generation}, including object discovery, segmentation, and predicate prediction.

\subsection{Foundation Models for Segmentation}

Foundation models such as SAM~\cite{kirillov2023segment} and SAM2~\cite{ravi2024sam} provide powerful engines for promptable segmentation and video mask propagation. SAM2, in particular, delivers high-quality temporal consistency. However, these models are class-agnostic and relation-agnostic: they cannot identify object categories, infer interactions, or discover interacting entities from a prompted subject.

Click2Graph fills this gap by introducing two components, the \emph{Dynamic Interaction Discovery Module} and the \emph{Semantic Classification Head}, that transform SAM2’s geometric outputs into pixel-accurate, temporally consistent scene graphs.

%% file: sec/3_method.tex
\section{Methodology}
\label{sec:Method}

\begin{figure*}[t]
    \centering
    \includegraphics[width=1\linewidth]{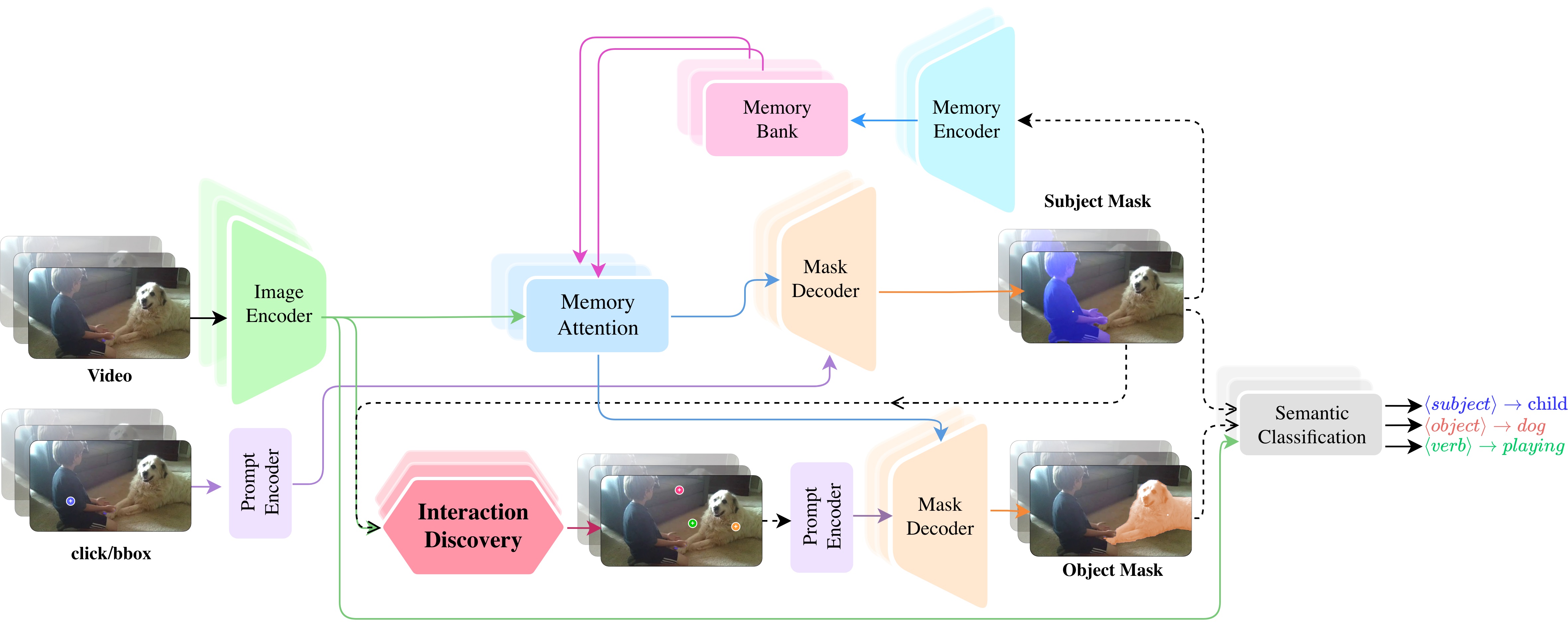}
    \caption{Overview of the \textbf{Click2Graph} architecture for user-guided Panoptic Video Scene Graph Generation. From a single user prompt, the system segments and tracks the subject, discovers interacting objects via the Dynamic Interaction Discovery Module (DIDM), and predicts subject--object--predicate triplets using the Semantic Classification Head (SCH).}
    \label{fig:tcdsg2-arch}
\end{figure*}

\subsection{Problem Formulation}

Given a video $\mathbf{V} = \{\mathbf{I}_1, \ldots, \mathbf{I}_T\}$ with $T$ frames and an initial user prompt $\mathbf{P}_i$ (a point, box, or mask) specifying the subject of interest, the goal of Click2Graph is to generate structured \emph{interaction tracklets}. Each tracklet describes a subject $s_i$, one of its interacting objects $o_{i,j}$, the relationship $r_{i,j}$ between them, and the corresponding panoptic masks over time.

Formally, for subject $i$, the set of interaction tracklets is:
\[
\mathbf{AT}_i = 
\left\{
    \mathbf{at}^i_j = 
    \langle 
        s_i,\,
        o_{i,j},\,
        r_{i,j},\,
        \mathbf{SM},\,
        \mathbf{OM},\,
        t_{\text{start}},\,
        t_{\text{end}}
    \rangle
\right\}_{j=1}^{M_i},
\]
where $\mathbf{SM}$ and $\mathbf{OM}$ denote the subject and object panoptic masks across the active temporal window $[t_{\text{start}}, t_{\text{end}}]$ and ${M_i}$ is the number of activities carried by subject ${s_i}$. Images are treated as a special case with $T=1$. The model supports multiple subjects, and users may introduce new prompts at any time.

\subsection{Network Architecture}

Click2Graph builds on \textbf{SAM2}~\cite{ravi2024sam2segmentimages}, a promptable video segmentation model that produces fine-grained, temporally consistent masks from sparse visual prompts. SAM2 is class-agnostic and yields masks for one object per prompt. To enable interaction discovery and semantic reasoning, we introduce two modules:

\begin{enumerate}
    \item \textbf{Dynamic Interaction Discovery Module (DIDM)} — predicts a set of subject-conditioned object prompts.
    \item \textbf{Semantic Classification Head (SCH)} — predicts subject, object, and predicate labels for discovered segments.
\end{enumerate}

Together, these modules transform SAM2 from a geometric segmentation backbone into a full panoptic video scene graph generator, see Figure~\ref{fig:tcdsg2-arch}, for a comprehensive overview of our architecture. 

\subsection{Dynamic Interaction Discovery Module}

It is a lightweight, set-based transformer module designed to convert a single user prompt into a fixed set of spatially precise object prompts. DIDM:
\begin{enumerate}
    \item Receives the encoded image features from the SAM2 backbone,
    \item For a given subject, generates a dedicated subject feature token by combining a learnable subject embedding with a feature vector derived from the subject's segmentation mask. This token is then prepended to a fixed set of $N_q$ learnable object query embeddings,
    \item Passes these combined query tokens through a series of Transformer layers, where they perform cross-attention against the image features. The queries are trained to shift their attention and encode the presence and location of objects interacting with the subject,
    \item Maps the refined object tokens to the normalized $(x, y)$ coordinates for the discovered interacting object prompts
\end{enumerate}

\begin{figure*}
    \centering
    \includegraphics[width=0.9\linewidth]{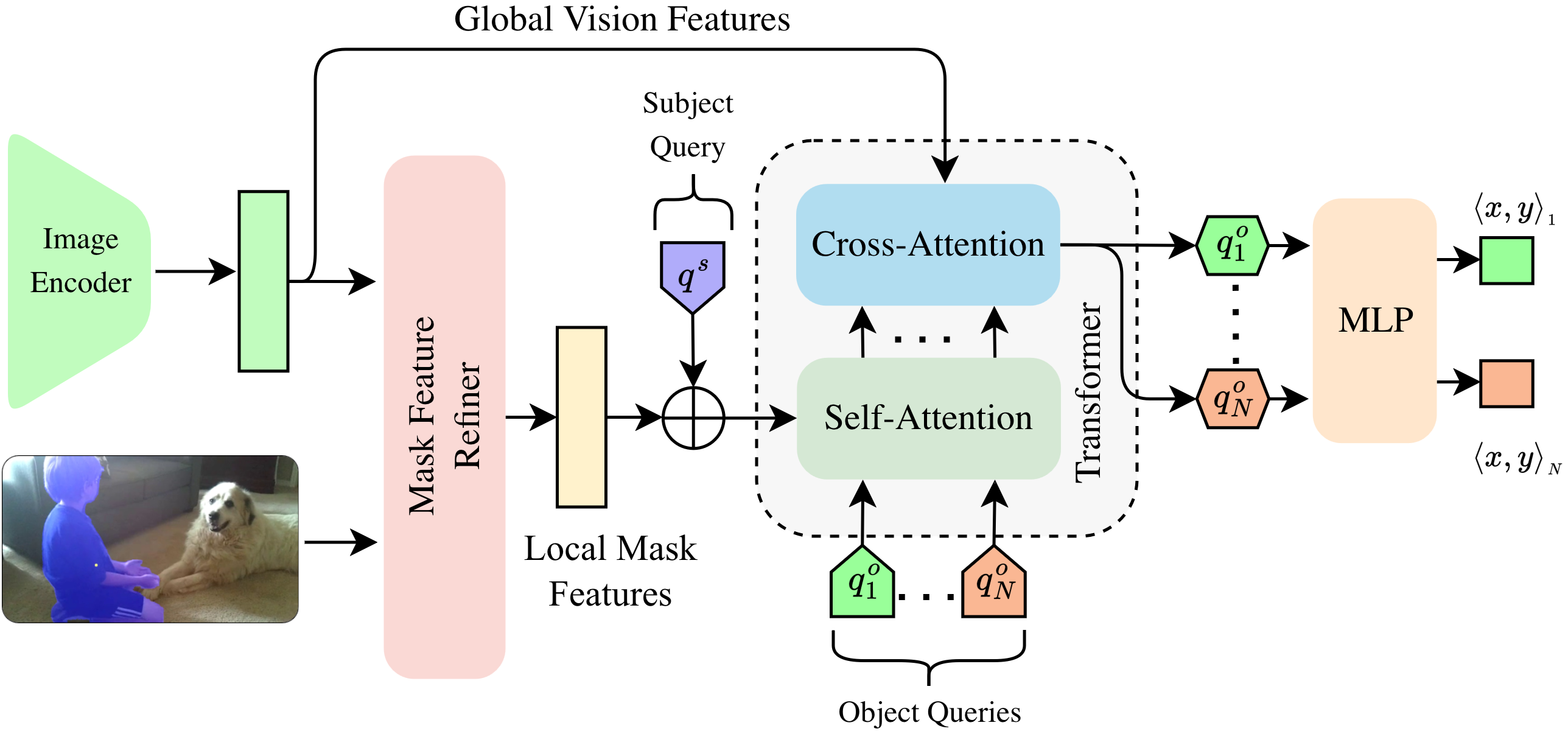}
    \caption{Architecture of the \textbf{Dynamic Interaction Discovery Module (DIDM)}. A single user-prompted subject prompt is transformed into $N_q$ predicted object prompts. It combines a feature vector derived from the subject mask with learnable object queries. These tokens pass through a Transformer decoder, which performs cross-attention over the image features, enabling the module to autonomously predict the precise locations (via the Point Prediction Head) of all entities interacting with the prompted subject.}
    \label{fig:prompt_generator}
\end{figure*}

The predicted points serve as prompt locations that SAM2 uses to segment candidate interacting objects. We empirically set $N_q = 3$ to exceed the typical number of objects interacting with a subject. Figure~\ref{fig:prompt_generator} illustrates the module's design.

\subsection{Semantic Classification Head}

This module bridges the gap between geometric outputs (masks) and structured, relational understanding (scene graphs). It performs the final semantic inference, classifying both objects and their relationships. The Semantic Classification Head:
\begin{enumerate}
    \item Extract semantic features by spatially aggregating the vision features (from the SAM2 encoder) over the predicted segmentation masks,
    \item Passes the aggregated subject and object features through a dedicated Multilayer Perceptron (MLP) to predict the subject's class label ($s_i$) and the object's class label ($o_{i,j}$) respectively,
    \item Concatenates the dedicated features from SAM2 Mask Decoder, specifically, the \textit{obj\_ptr} query token for the subject and the discovered object to form a subject-object pair representation, 
    \item Passes this joint feature vector through a separate MLP to predict the complex relationship predicate ($r_{i,j}$) 
\end{enumerate}

For each prompted subject $i$, the output is:
\begin{equation}
    O(\mathbf{I}_t \mid P_i) = \bigcup_{j=1}^{N_q} 
    \langle s_i,\; o_{i,j},\; r_{i,j} \rangle.
    \label{eq:output}
\end{equation}

\subsection{Training Objective}

We formulate our objective as a strategically composed multi-task loss to effectively optimize the heterogeneous output types of our framework: panoptic segmentation masks, precise control over object discovery, and structured semantic reasoning:

\[
\mathcal{L}_{\text{total}}
=
\mathcal{L}_{\text{mask}}
+
\mathcal{L}_{L2}
+
\mathcal{L}_{\text{sub}}
+
\mathcal{L}_{\text{obj}}
+
\mathcal{L}_{\text{rel}}.
\]

\paragraph{Mask Loss.}  
For both subject and discovered object masks, we use a combination of:
\[
\mathcal{L}_{\text{mask}}
=
\mathcal{L}_{\text{BCE}}
+
\mathcal{L}_{\text{IoU}}
+
\mathcal{L}_{\text{Dice}}.
\]

\paragraph{Prompt Localization Loss.}  
Each DIDM-predicted point is supervised with:
\[
\mathcal{L}_{L2}
=
\|\hat{p} - p^\ast\|_{2}^{2},
\]
where $p^\ast$ is a ground-truth object interior point.

\paragraph{Semantic Prediction Loss.}  
We apply cross-entropy losses to:
$
\mathcal{L}_{\text{sub}}, \quad
\mathcal{L}_{\text{obj}}, \quad
\mathcal{L}_{\text{rel}}.
$

\paragraph{Set-Based Hungarian Matching.}  
Because DIDM and SCH generate a fixed set of predictions, we adopt the bipartite matching strategy from DETR~\cite{carion2020end} to align predictions with ground-truth interaction sets.

\subsection{Training and Inference Details}

We use SAM2.1-Large as the backbone and freeze its 224M parameters. DIDM and SCH introduce approximately 5M trainable parameters. Training uses AdamW with learning rate $5{\times}10^{-4}$ for SCH parameters and a cosine annealing schedule with start value $5{\times}10^{-5}$ and end value $1{\times}10^{-5}$  for DIDM parameters. We train for 400 epochs, sampling 8-frame clips per batch following SAM2’s video-centric strategy.

For each loss term $\mathcal{L}_l$ we use an appropriate loss weight $\lambda_l$ which are set as:
\[
\begin{aligned}
\lambda_{\text{BCE}} &= 10,\;
\lambda_{\text{Dice}} = 1,\;
\lambda_{\text{IoU}} = 1, \\[2pt]
\lambda_{L2} &= 20,\;
\lambda_{\text{sub}} = 10,\;
\lambda_{\text{obj}} = 10,\;
\lambda_{\text{rel}} = 20.
\end{aligned}
\]

Inference runs at $\sim 10$ FPS on an NVIDIA A100 (40GB), with a memory footprint of $\sim 7$GB for a Video input resolution of (1024x1024).

\subsection{Ground-Truth Point Sampling Strategy}
\label{subsec:point_sample}

Training DIDM requires stable ground-truth points inside each object mask. Boundary points are ambiguous for promptable models like SAM2, whereas interior points yield clearer supervisory signals.

We therefore:
\begin{enumerate}
    \item Compute the distance transform of each object mask,
    \item Assign each pixel a sampling probability proportional to its distance from the mask boundary,
    \item Sample core interior points as high-quality targets.
\end{enumerate}

This distance-weighted sampling generates robust supervision for DIDM’s point regression and improves object discovery accuracy.

%% file: sec/4_dataset.tex
\section{Dataset: OpenPVSG}

The Panoptic Video Scene Graph (PVSG) task requires pixel-level grounding of entities and their relationships across time. We evaluate Click2Graph on the \textbf{Open Panoptic Video Scene Graph (OpenPVSG)} dataset introduced by Yang et al.~\cite{yang2023pvsg}, which provides the most comprehensive benchmark for panoptic-level video scene graph generation.

\paragraph{Scale and Composition:}
OpenPVSG contains \textbf{400} videos totaling approximately \textbf{150k} frames at 5 FPS. On those videos, each subject is typically interacting with $\leq 2$ objects per frame. The data spans a wide range of environments and camera styles, aggregated from:
\begin{itemize}
    \item VidOR~\cite{shang2019annotating} (289 videos),
    \item EPIC-Kitchens~\cite{Damen2018EPICKITCHENS} (55 videos),
    \item Ego4D~\cite{grauman2022ego4d} (56 videos).
\end{itemize}
The dataset includes both third-person and egocentric perspectives, enabling evaluation under diverse motion patterns, object configurations, and interaction types.

\paragraph{Annotations:}
The annotation set includes:
\begin{itemize}
    \item \textbf{126 object categories} grounded with pixel-accurate panoptic segmentation,
    \item \textbf{57 relationship predicates} covering spatial, contact, and interaction types,
    \item Temporally consistent instance masks and relation trajectories.
\end{itemize}
Panoptic masks allow detailed grounding of both “things’’ and “stuff’’ classes, which is essential for modeling non-rigid entities and background interactions.

\paragraph{Relevance to Click2Graph:}
OpenPVSG provides a challenging testbed for user-guided PVSG for three reasons:
\begin{enumerate}
    \item \textbf{High visual and temporal diversity:} videos include complex camera motion, occlusions, multiple interacting entities, and indoor/outdoor environments.
    \item \textbf{Fine-grained semantic space:} the large number of closely related object and predicate classes exposes the difficulty of semantic reasoning.
    \item \textbf{Panoptic-level grounding:} pixel-accurate masks are necessary for evaluating prompt localization, segmentation, and relationship prediction.
\end{enumerate}

These characteristics make OpenPVSG an ideal benchmark for assessing Click2Graph’s ability to combine visual prompting, object discovery, panoptic segmentation, and relational reasoning in a unified framework.

%% file: sec/5_evalmetrics.tex
\section{Evaluation Metrics}
\label{sec:metrics}

Click2Graph integrates visual prompting, interaction discovery, panoptic segmentation, and semantic reasoning into a unified pipeline. Standard SGG metrics such as Predicate Classification (PREDCLS), Scene Graph Classification (SGCLS), and Scene Graph Detection (SGDET) are therefore inappropriate to characterize system performance. We evaluate using three complementary recall-based metrics that provides a fine-grained evaluation of our model’s spatial precision, prompt generation reliability, and overall scene graph accuracy.

\vspace{0.5em}
\noindent\textbf{1.\;Recall@K (End-to-End Semantic Interaction Recall)}  
Recall@K (R@K) measures full triplet correctness. A prediction  
$\langle s_i, o_{i,j}, r_{i,j} \rangle$ is counted as correct if:
\begin{enumerate}
    \item Subject, object, and predicate labels match the ground truth; and
    \item The predicted subject and object masks both achieve $\mathrm{IoU} \ge \tau$ with the corresponding ground-truth masks.
\end{enumerate}
Predictions are ranked by confidence, and only the top-$K$ are used. This metric evaluates the complete Click2Graph pipeline, combining DIDM, SAM2 segmentation, and SCH semantic reasoning. Following prior PVSG work, we set the IoU threshold to $\tau = 0.5$.

\vspace{0.5em}
\noindent\textbf{2.\;Spatial Interaction Recall (SpIR)}  
SpIR isolates the quality of spatial grounding. While calculating this metric, a subject–object pair is considered correct if it satisfies the following requirement:
\[
\mathrm{IoU}(\hat{\mathbf{SM}}, \mathbf{SM}^\ast) \ge \tau
\quad\text{and}\quad
\mathrm{IoU}(\hat{\mathbf{OM}}, \mathbf{OM}^\ast) \ge \tau,
\]
regardless of predicted class or predicate labels.  
This metric evaluates the combined effectiveness of DIDM in producing appropriate object prompts and SAM2 in propagating precise panoptic masks over time.

\vspace{0.5em}
\noindent\textbf{3.\;Prompt Localization Recall (PLR)}  
PLR measures the accuracy of DIDM’s predicted object prompt points. A discovered object prompt $\hat{p}_{i,j}$ is counted as correct if it lies within the ground-truth object mask:
\[
\hat{p}_{i,j} \in \mathbf{OM}_{i,j}^\ast.
\]
PLR thus assesses the reliability of interaction discovery independently of subsequent segmentation or semantic prediction.

\vspace{0.5em}
\noindent\textbf{Evaluation Protocol.}  
Because prompt-based systems are sensitive to initial user inputs, we evaluate robustness by repeating each experiment \textbf{25 times}, sampling a unique initial point from the subject’s ground-truth mask for each run. We report all metrics as \textbf{mean}~$\pm$~\textbf{standard deviation} across runs. 

\vspace{0.5em}

\begin{table}[t]
\centering
\setlength{\tabcolsep}{5pt}
\caption{Comparison of standard Recall@K metrics between Click2Graph and prior automated PVSG approaches. Prior methods generate full-frame proposals and must detect subjects; Click2Graph receives a subject prompt, reflecting its interactive setting.}
\begin{tabular}{lcc}
\toprule
\textbf{Method} & \textbf{Recall@3} & \textbf{Recall@20} \\
\midrule
PVSG \cite{yang2023pvsg} + IPS+T \cite{cheng2022masked, wang2021different} & - & 3.88 \\
PVSG \cite{yang2023pvsg} + VPS \cite{cheng2022masked, li2022video} & - & 0.42 \\
MACL \cite{nguyen2025motion} + IPS+T & - & \textbf{4.51} \\
MACL \cite{nguyen2025motion} + VPS & - & 0.84 \\
\textbf{Click2Graph (Ours)} & \textbf{2.23} & - \\
\bottomrule
\end{tabular}
\label{tab:sir_results}
\end{table}

%% file: sec/6_result.tex
\section{Results \& Ablations}
We evaluate Click2Graph on the OpenPVSG benchmark using the three metrics introduced in Section~\ref{sec:metrics}. These metrics allow us to separately assess (1) semantic triplet reasoning, (2) segmentation and interaction grounding quality, and (3) the reliability of object prompt generation. 

\paragraph{End-to-End Performance:}
Table~\ref{tab:sir_results} compares Click2Graph with prior automated PVSG approaches. Unlike these methods, which generate dense full-frame proposals and must detect subjects, our work receives a subject prompt and produces only the interaction-centric predictions associated with that target. Despite generating far fewer predictions per frame ($N_q=3$, compared to $\sim$100 in automated baselines), Click2Graph achieves competitive R@K scores. This demonstrates that targeted, user-guided reasoning can reduce the search space while preserving strong semantic alignment. Furthermore, the interactive paradigm makes Click2Graph complementary to fully automated PVSG methods, offering a practical path toward controllable and corrective scene graph generation.

\paragraph{Robustness to Prompt Type:}
We study how the quality of user-specified prompts influences Click2Graph by comparing three forms of input: a single point, a bounding box, and a full segmentation mask. During training, point and box prompts are sampled with high probability (0.49 each), reflecting low-effort user inputs, while mask prompts are used rarely (0.02). As shown in Table~\ref{tab:frame_avg_results}, performance varies modestly across prompt types: masks yield slightly higher scores, as expected, but \emph{all three} provide stable results, with low variance across runs. This confirms that Click2Graph is robust to imperfect or low-precision user interactions, which is a key requirement for practical deployment.

\begin{table}[t]
\centering
\setlength{\tabcolsep}{3pt}
\renewcommand{\arraystretch}{0.9}
\caption{Ablation experiment showing robustness to different prompt types.}
\begin{tabular}{llccc}
\toprule
\textbf{Dataset} & \textbf{Prompt} & \textbf{R@3} & \textbf{SpIR} & \textbf{PLR} \\
\midrule
Epic K. 
& Mask & 1.78 & 24.22 & 30.67 \\
& Point & 1.14\footnotesize{$\pm$0.38} & 23.04\footnotesize{$\pm$1.08} & \textbf{32.06\footnotesize{$\pm$0.81}} \\
& BBox & \textbf{2.08\footnotesize{$\pm$0.06}} & \textbf{25.02\footnotesize{$\pm$0.09}} & 31.96\footnotesize{$\pm$0.09} \\
\midrule
Ego4d 
& Mask & \textbf{0.73} & 17.22 & 38.37 \\
& Point & 0.56\footnotesize{$\pm$0.04} & 16.21\footnotesize{$\pm$1.04} & \textbf{39.87\footnotesize{$\pm$0.38}} \\
& BBox & 0.72\footnotesize{$\pm$0.06} & \textbf{17.49\footnotesize{$\pm$0.32}} & 38.97\footnotesize{$\pm$0.11} \\
\midrule
Vidor 
& Mask & \textbf{3.33} & \textbf{18.77} & \textbf{30.82} \\
& Point & 2.72\footnotesize{$\pm$0.25} & 15.37\footnotesize{$\pm$0.55} & 28.86\footnotesize{$\pm$0.34} \\
& BBox & 3.18\footnotesize{$\pm$0.10} & 17.59\footnotesize{$\pm$0.36} & 30.13\footnotesize{$\pm$0.23} \\
\bottomrule
\end{tabular}
\label{tab:frame_avg_results}
\end{table}

\paragraph{Contribution of System Components:}
The three metrics jointly reveal the behavior of Click2Graph's submodules. High PLR scores indicate that the Dynamic Interaction Discovery Module reliably generates subject-conditioned prompts that fall inside the correct object regions. Strong SpIR performance demonstrates that SAM2, when guided by these prompts, yields accurate panoptic masks for both subjects and interacting objects. R@K remains the most challenging metric, reflecting the difficulty of fine-grained label and predicate classification. Most semantic errors arise from confusions between visually similar categories (e.g., \textbf{child} vs.~\textbf{baby}, \textbf{box} vs.~\textbf{bag}, \textbf{floor} vs.~\textbf{ground}), consistent with the long-tail and high redundancy of the OpenPVSG semantic space.

\begin{figure*}[h]
    \centering

    \begin{subfigure}{0.3\textwidth}
        \centering
        \includegraphics[width=\linewidth]{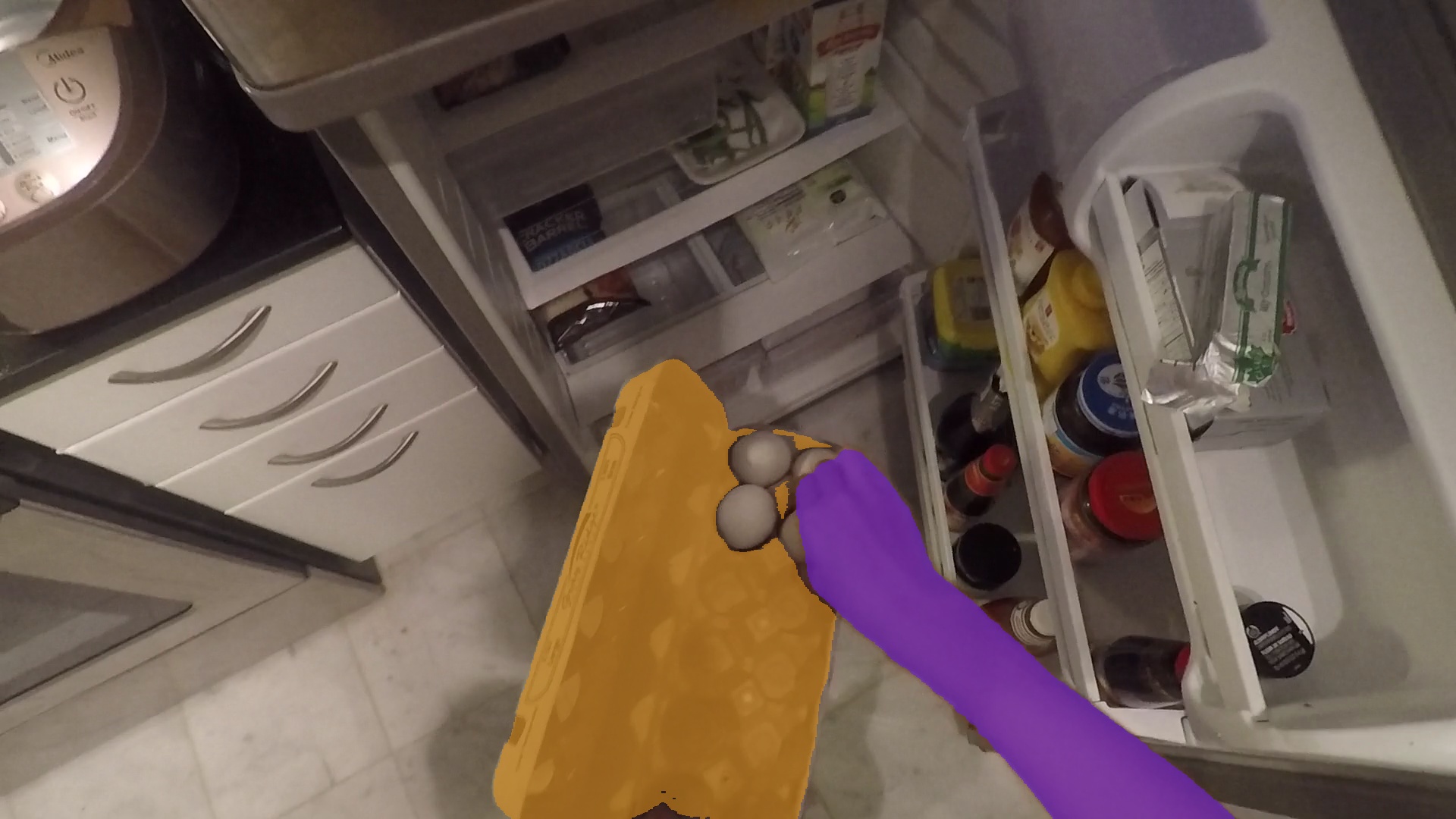}
        \caption{\parbox{\linewidth}{\centering
        Prediction: \colorTrip{adult}{box}{holding} \\
        GT: \colorTrip{adult}{box}{holding}}}
    \end{subfigure}
    \begin{subfigure}{0.3\textwidth}
        \centering
        \includegraphics[width=\linewidth]{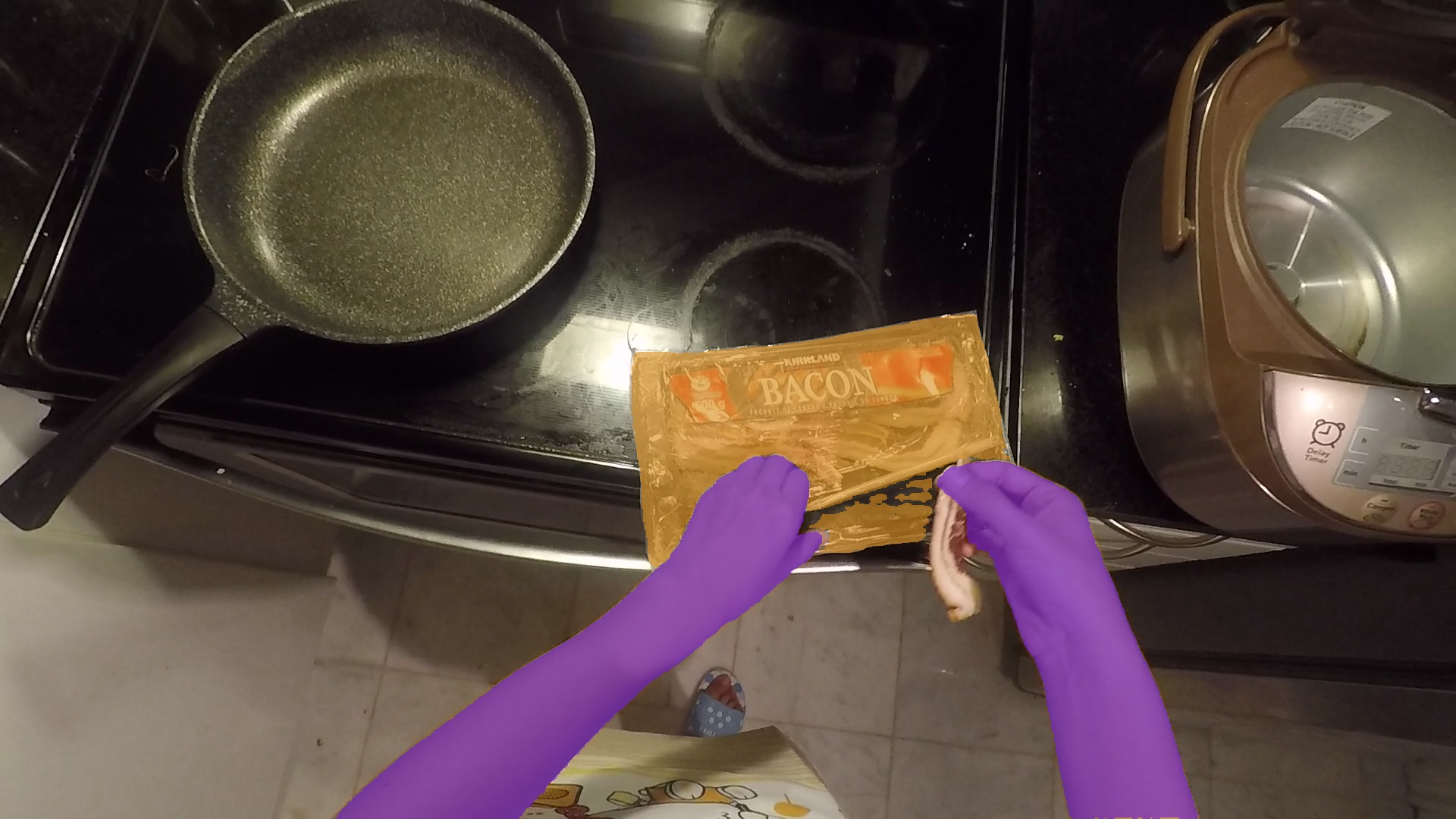}
        \caption{\parbox{\linewidth}{\centering
        Prediction: \colorTrip{adult}{bag}{holding} \\
        GT: \colorTrip{adult}{bag}{holding}}}
    \end{subfigure}
    \begin{subfigure}{0.3\textwidth}
        \centering
        \includegraphics[width=\linewidth]{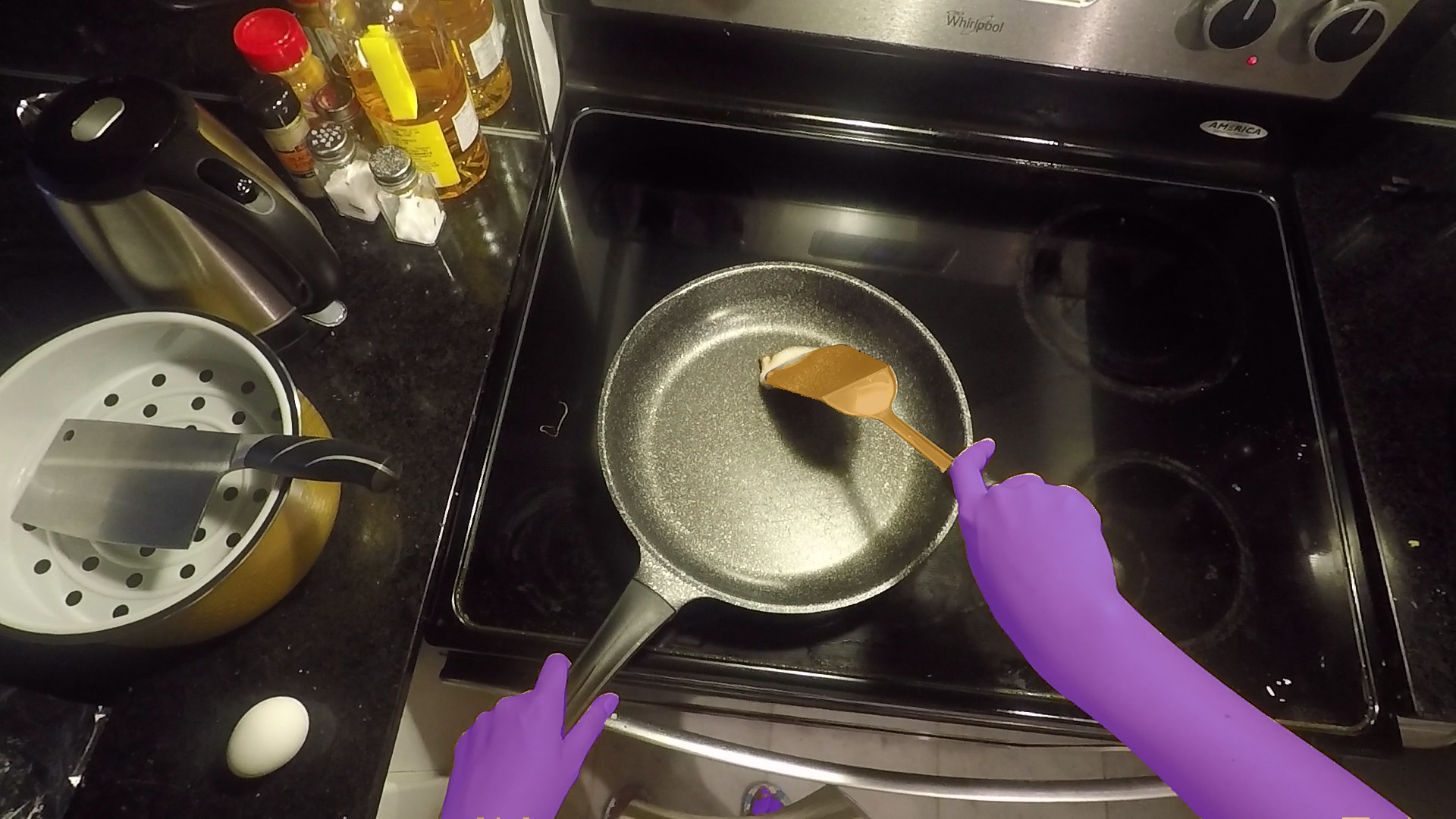}
        \caption{\parbox{\linewidth}{\centering
        Prediction: \colorTrip{adult}{spatula}{holding} \\
        GT: \colorTrip{adult}{spatula}{holding}}}
    \end{subfigure}

    \vskip\baselineskip

    \begin{subfigure}{0.3\textwidth}
        \centering
        \includegraphics[width=\linewidth]{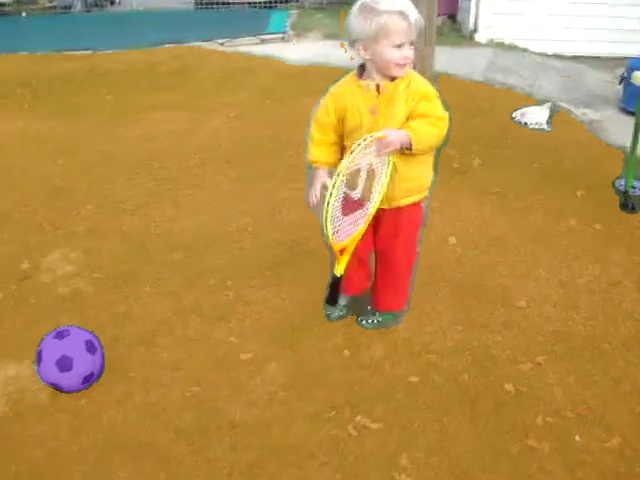}
        \caption{\parbox{\linewidth}{\centering
        Prediction: \colorTrip{ball}{grass}{on} \\
        GT: \colorTrip{ball}{grass}{on}}}
    \end{subfigure}
    \begin{subfigure}{0.3\textwidth}
        \centering
        \includegraphics[width=\linewidth]{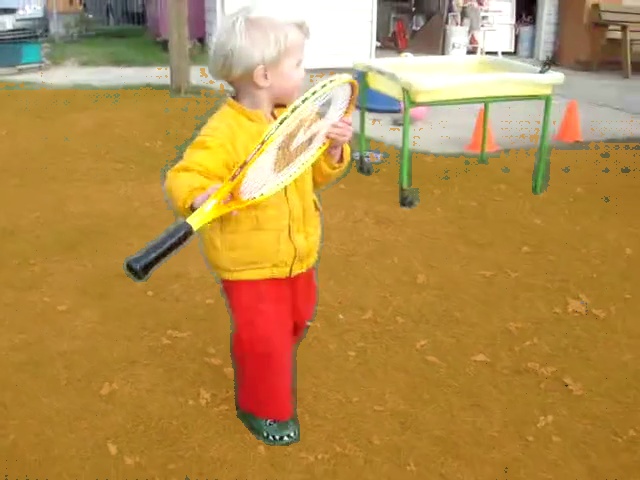}
        \caption{\parbox{\linewidth}{\centering
        Subject temporarily occluded by camera motion}}
    \end{subfigure}
    \begin{subfigure}{0.3\textwidth}
        \centering
        \includegraphics[width=\linewidth]{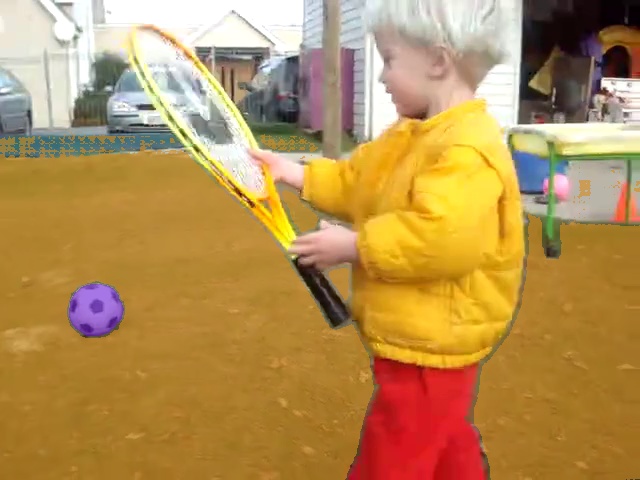}
        \caption{\parbox{\linewidth}{\centering
        Prediction: \colorTrip{ball}{grass}{on} \\
        GT: \colorTrip{ball}{grass}{on}}}
    \end{subfigure}

    \vskip\baselineskip

    \begin{subfigure}{0.3\textwidth}
        \centering
        \includegraphics[width=\linewidth]{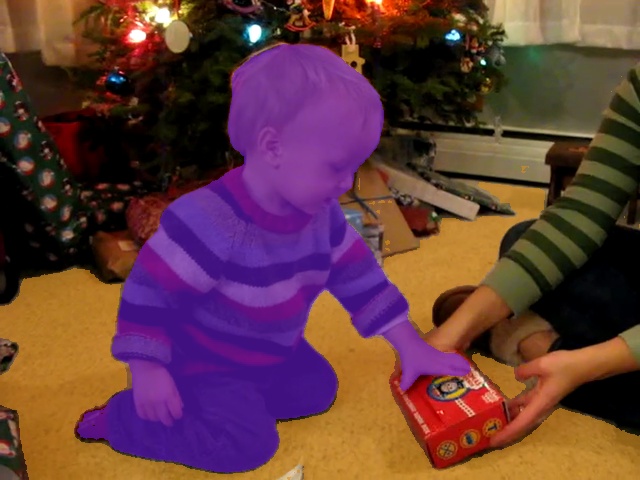}
        \caption{\parbox{\linewidth}{\centering
        Prediction: \colorTrip{child}{floor}{sitting} \\
        GT: \colorTrip{child}{floor}{on}}}
    \end{subfigure}
    \begin{subfigure}{0.3\textwidth}
        \centering
        \includegraphics[width=\linewidth]{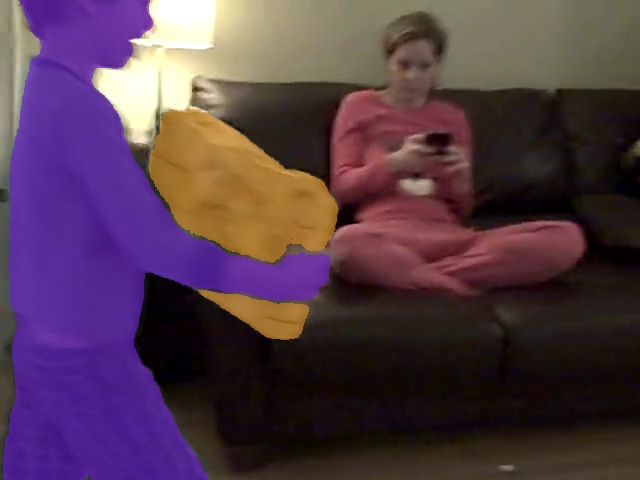}
        \caption{\parbox{\linewidth}{\centering
        Prediction: \colorTrip{child}{box}{holding} \\
        GT: \colorTrip{child}{gift}{holding}}}
    \end{subfigure}
    \begin{subfigure}{0.3\textwidth}
        \centering
        \includegraphics[width=\linewidth]{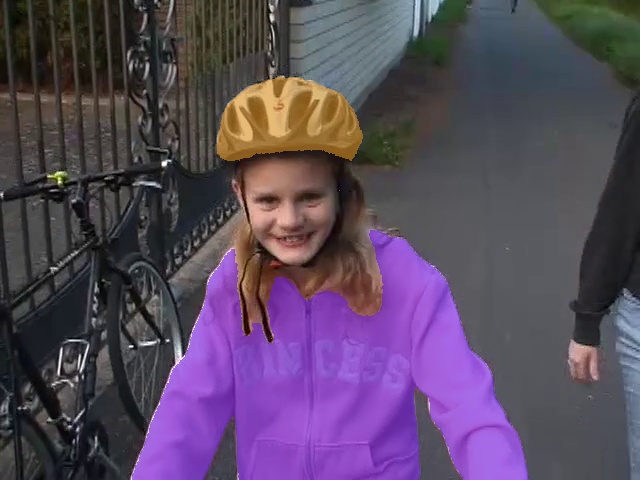}
        \caption{\parbox{\linewidth}{\centering
        Prediction: \colorTrip{child}{helmet}{wearing} \\
        Triplet not in Ground Truth}}
    \end{subfigure}

    \caption{Qualitative results illustrating correct predictions, occlusion robustness, and typical failure cases.}
    \label{fig:qualitative}
\end{figure*}

\paragraph{Importance of DIDM:}
To isolate the contribution of the Dynamic Interaction Discovery Module, we replace it with a heuristic that samples prompts from a dataset-level object-probability heatmap. The heuristic assigns high likelihood to locations where objects commonly appear but is not conditioned on the prompted subject. Table~\ref{tab:ab_disc} shows that this replacement severely degrades PLR, SpIR, and R@K across all datasets. This highlights that subject-conditioned prompt generation is essential for interaction-centric reasoning—generic object priors are insufficient to capture the relational structure required for PVSG.

\begin{table}[h]
\centering
\caption{Comparison of the different metric based on the interaction discovery strategy.}
\begin{tabular}{llccc}
\toprule
\textbf{Dataset} & \textbf{Strategy} &
\multicolumn{3}{c}{\textbf{Metric}} \\
\cmidrule(lr){3-5}
& & \textbf{R@3} & \textbf{SpIR} & \textbf{PLR} \\
\midrule

\textbf{Epic K.} 
& Heuristic  & 0.62 & 5.14 & 10.60 \\
& DIDM(ours) & \textbf{2.08} & \textbf{25.02} & \textbf{32.06} \\
\midrule

\textbf{Ego4d} 
& Heuristic  & 0.28 & 4.26 & 9.30 \\
& DIDM(ours) & \textbf{0.73} & \textbf{17.49} & \textbf{39.87} \\
\midrule

\textbf{Vidor} 
& Heuristic  & 0.68 & 4.66 & 10.19 \\
& DIDM(ours) & \textbf{3.33} & \textbf{18.77} & \textbf{30.82} \\
\bottomrule
\end{tabular}
\label{tab:ab_disc}
\end{table}

\paragraph{Qualitative Analysis:}
Figure~\ref{fig:qualitative} illustrates Click2Graph’s behavior across diverse scenarios. In the first row, the system correctly recovers multiple interacting objects and produces coherent triplets. The second row demonstrates temporal robustness: even after partial occlusion or momentary subject disappearance, the system continues to produce consistent predictions. Failure cases (third row) typically involve predicate granularity (\textbf{on} vs. \textbf{sitting}) or object categories with subtle visual differences (\textbf{gift} vs.~\textbf{box}). These examples visually corroborate our quantitative findings: segmentation and interaction discovery are reliable, while semantic classification remains the primary bottleneck.

%% file: sec/7_Conc_future.tex
\section{Conclusions and Future Work}

We introduced \textbf{Click2Graph}, the first user-guided framework for Panoptic Video Scene Graph Generation. By combining a single visual prompt with subject-conditioned interaction discovery and semantic reasoning, Click2Graph enables controllable, interpretable video understanding. Central to the system is the Dynamic Interaction Discovery Module, which reliably generates object prompts conditioned on the user-specified subject, and the Semantic Classification Head, which elevates promptable segmentation into full triplet prediction. Together, these components transform SAM2 into a complete PVSG pipeline capable of structured, interaction-centric reasoning.

Experiments on the OpenPVSG benchmark demonstrate that Click2Graph achieves strong spatial grounding and reliable object discovery, while highlighting the challenges of fine-grained semantic classification in a large, diverse label space. Most errors arise from distinctions between visually similar object categories or predicates, suggesting that semantic reasoning, rather than segmentation or interaction discovery, is the primary bottleneck.

A limitation of the current system is that real-time user intervention is restricted to segmentation correction; users cannot directly modify predicted labels during inference, and such corrections do not yet feed back into the model. As future work, we plan to integrate a lightweight feedback mechanism in which user-provided label corrections dynamically update a set of learnable class embeddings. This would enable Click2Graph to adapt its semantic predictions over time and maintain consistency across future frames.

Beyond label correction, Click2Graph opens several promising research directions, including (1) integrating language models to enhance predicate reasoning and reduce fine-grained semantic confusion, (2) developing multi-subject prompting strategies for complex multi-agent interactions, and (3) leveraging interactive supervision to improve long-tail predicate learning. By unifying promptable segmentation with subject-conditioned relational inference, Click2Graph offers a foundation for the next generation of interactive, human-centered video scene understanding systems.